\documentclass[runningheads]{llncs}
\usepackage[T1]{fontenc}
\usepackage{graphicx}
\usepackage{placeins}
\usepackage{booktabs}
\usepackage{graphicx} % Required for inserting images
\usepackage{amsmath}
\usepackage{subcaption}
\usepackage{indentfirst}
\usepackage{tikz}
\usepackage{amsfonts}
\usepackage{algorithm}      % 提供algorithm浮动体环境
\usepackage{algorithmic}  % 提供algorithmic环境（注意：较新版本使用algpseudocode而非algorithmic）
\usepackage{amsmath}        % 提供数学符号支持
\begin{document}
\title{Bayesian
Optimization for Enhanced Language Models: Optimizing Acquisition Functions}
\titlerunning{Bayesian Optimization for Language Models}
\author{
Zishuo Bao\textsuperscript{1,†}, 
Yibo Liu\textsuperscript{2,†}, 
Changyutao Qiu\textsuperscript{3,†} 
\thanks{†These authors contributed equally to this work. The authors are listed alphabetically by surname.} 
}
\institute{
Fuzhou University, Fuzhou ,China\and
Xi'an Jiaotong-Liverpool University, Suzhou ,China    \and
Dalian University of Technology, Dalian, China 
}

\maketitle              % typeset the header of the contribution

\begin{abstract}

With the rise of different language model architecture, fine-tuning is becoming even more important for down stream tasks Model gets messy, finding proper hyperparameters for fine-tuning. Although BO has been tried for hyperparameter tuning, most of the existing methods are oblivious to the fact that BO relies on careful choices of acquisition functions, which are essential components of BO that guide how much to explore versus exploit during the optimization process; Different acquisition functions have different levels of sensitivity towards training loss and validation performance; existing methods often just apply an acquisition function no matter if the training and validation performance are sensitive to the acquisition function or not. This work introduces{Bilevel - BO - SWA}, a model fusion approach coupled with a bilevel BO strategy to improve the fine - tunning of large language models. Our work on mixture of acquisition functions like EI and UCB into nested opt loops, where inner loop perform minimization of training loss while outer loops optimized w.r.t. val metric. Experiments on GLUE tasks using RoBERTA - base show that when using EI and UCB, there is an improvement in generalization, and fine - tuning can be improved by up to 2.7%.


\end{abstract}
\section{Introduction}
\noindent The advent of Transformer-based language models, such as BERT \cite{devlin2018bertbaseline}, GPT \cite{radford2018improving}, and RoBERTa \cite{liu2019roberta}, has revolutionized natural language processing (NLP) by enabling transfer learning through large-scale pretraining. These models capture rich linguistic patterns from unlabeled corpora, which can then be adapted to downstream tasks via fine-tuning—a process that adjusts pretrained weights on task-specific labeled data \cite{howard2018universal}. 

Fine-tuning Transformer-based language models on downstream datasets has become a standard approach\cite{devlin2018bertbaseline}. However, the performance of fine-tuned models heavily depends on hyperparameter selection\cite{feurer2019hyperparameter}, making optimization a critical yet underexplored challenge. Carefully tuning of hyperparameters, including learning rate and weight decay is crucial in optimizing \cite{feurer2019hyperparameter}. While Bayesian optimization (BO) has emerged as a powerful tool for hyperparameter tuning \cite{snoek2012practical}, most existing work relies on a single acquisition function without considering their distinct behaviors. Minimizing loss and boosting accuracy may require different degrees of exploration. UCB often reacts more strongly to training loss, while EI focuses on maximizing accuracy. We hypothesize that mixing these functions at different stages could have more balanced results, but no prior work examines this. Our experiments fill this gap, exploring how pairing EI and UCB across nested loops can optimize both training dynamics and final performance for large language models.

Model fusion offers a promising alternative to traditional fine-tuning by aggregating multiple models into a single one, reducing the need for ensembles \cite{model_fusion}. While techniques like Stochastic Weight Averaging (SWA) have shown success in improving generalization in computer vision, their application to NLP is less effective due to the misalignment between loss functions and evaluation metrics, resulting in suboptimal performance.

In this work, {Bilevel-BO-SWA}, a model fusion approach that addresses these challenges. Our method uses {Bilevel Bayesian Optimization (BO)}\cite{bo_llm,snoek2012practical}, focusing specifically on the optimization of acquisition functions, which play a crucial role in guiding the search process. We also explore combinations of {Expected Improvement (EI)} and {Upper Confidence Bound (UCB)} acquisition functions, both independently and in different configurations. We examine scenarios where EI is applied in the outer optimization layer and UCB in the inner layer, as well as the reverse configuration. Our goal is to determine how different pairings of acquisition functions impact the identification of optimal model fusion parameters.

Our experiments demonstrate that the selection and arrangement of acquisition functions significantly influence model performance, with tailored strategies leading to notable improvements over existing fusion techniques. Specifically, we evaluate our approach on multiple NLP tasks using RoBERTa and compare it against the GLUE benchmark\cite{wang2019glue}. The results show that {EI-UCB} achieves the highest average score (76.82), outperforming the standard fine tuning by 2.7\%. These findings highlight the critical role of acquisition functions in improving the generalization and optimization of large language models.

\noindent \textbf{To summarize, this work makes three key contributions:}\par
\vspace{0.5em}
\noindent 1. We propose \textsc{Bilevel-BO-SWA}, the first framework that integrates bilevel Bayesian optimization with model fusion for language model fine-tuning, providing a principled approach to hyperparameter optimization.\par
\vspace{0.5em} 
\noindent 2. We develop a novel strategy for combining EI and UCB acquisition functions in hierarchical optimization loops, theoretically analyzing and empirically validating their complementary effects.\par
\vspace{0.5em}
\noindent 3. Our extensive experiments on the GLUE benchmark demonstrate consistent improvements, with the EI-UCB configuration achieving 2.7\% higher accuracy than conventional fine-tuning.

\section{Related Work}
\noindent This section reviews relevant research from three perspectives: language models, fine-tuning techniques, and Bayesian optimization.

\subsection{Language Models}

\noindent Recent years have witnessed significant progress in transformer-based language models such as BERT \cite{devlin2019bert}, RoBERTa \cite{liu2019roberta}, and GPT \cite{radford2018improving}. These models achieve state-of-the-art performance on various downstream natural language processing tasks by leveraging large-scale pretraining followed by task-specific fine-tuning. Despite their effectiveness, the high computational cost of training and the risk of overfitting remain major challenges, particularly when hyperparameters like learning rate, batch size, and weight decay are not well-tuned.

\subsection{Fine-tuning Techniques}

\noindent Fine-tuning pretrained language models is a common practice to adapt them to target tasks, but it often requires careful hyperparameter tuning to avoid overfitting and suboptimal performance. Model fusion methods, which aggregate predictions or weights from multiple models or training runs, have shown promise in stabilizing and improving model performance. One notable technique, Stochastic Weight Averaging (SWA) \cite{izmailov2018swa}, averages model weights across different checkpoints and has proven effective in computer vision benchmarks. However, the direct application of SWA and similar fusion methods to NLP is less successful, mainly due to differences in loss functions like cross-entropy in NLP versus softmax loss used in vision tasks \cite{seal}. This highlights the need for tailored fine-tuning and model fusion strategies in language model adaptation.

\subsection{Bayesian Optimization}
\noindent Bayesian Optimization (BO) has become a popular and effective framework for hyperparameter tuning in machine learning \cite{snoek2012practical}. BO provides a probabilistic model to guide the search for optimal hyperparameters by balancing exploration and exploitation of the search space. It has been successfully applied to a variety of machine learning models, including neural networks, to better optimize hyperparameters such as learning rates and regularization parameters. RaditionalBO relies on acquisition functions such as Expected Improvement (EI) and Upper Confidence Bound (UCB), which help determine the next set of hyperparameters to evaluate based on past evaluations\cite{shahriari2016taking}.

Standard BO techniques face challenges in high-dimensional search spaces and complex objective functions. In particular, can be challenging in multi-objective optimization scenarios\cite{morbo}. Recent work has addressed this issue by introducing hierarchical or bilevel optimization approaches, where the outer loop optimizes 
ion performance, and the inner loop focuses on loss \cite{bilevel_intro}. These approaches are useful for fine-tuning large models, optimizing both the architecture and the hyperparameters to improve performance\cite{falkner2018bohb}.

Compared with Jang et al.\cite{model_fusion}, who use a  multi-objective Bayesian Optimization, we introduce a bilevel approach that separates training loss optimization in the inner loop from outer loop. In addition, we explore pairing different acquisition functions across these two loops, to enable more flexible hyperparameter navigation and improve generalization. Our approach combines model fusion with Bilevel Bayesian optimization, using acquisition functions ( EI and UCB) with the bilevel BO framework. The advantage is guiding the search while balancing exploration and exploitation. Experiments are on different configurations of acquisition functions, where the outer layer optimizes for validation performance while the inner layer optimizes for training loss. Results show that the interaction between acquisition functions in the bilevel optimization process makes improvements.

\section{Methodology}

\noindent Our approach combines Bayesian Optimization (BO) with bilevel optimization to address the challenge of efficiently fine-tuning large language models. The methodology consists of three key components: (1) Expected Improvement for outer-loop optimization, (2) Upper Confidence Bound for inner-loop exploration, and (3) a bilevel framework that coordinates these acquisition functions in a hierarchical structure. This integrated approach enables systematic navigation of the hyperparameter space while balancing computational efficiency with model performance.

We first introduce the two acquisition functions that form the building blocks of our optimization strategy, then present the bilevel architecture that orchestrates their interaction. The mathematical formulations and implementation details are provided for each component, along with justification for their specific roles in our framework.

\subsection{Expected Improvement (EI)}

\noindent Expected Improvement (EI) is a widely used acquisition function in Bayesian Optimization (BO) that balance exploration and exploitation\cite{bo_llm}.  The core idea behind EI is to compute the expected improvement in the objective function value at a given point based on the posterior distribution of the objective function.

Mathematically, the expected improvement at a point \( \theta \) is defined as:
\[
EI(\theta) = \mathbb{E}[\max(f(\theta) - f_{\text{best}}, 0)],
\]
where \( f_{\text{best}} \) represents the current best observed value of the objective function, and \( f(\theta) \) is the predicted value of the objective at point \( \theta \). The expectation is taken with respect to the distribution of \( f(\theta) \), which is typically modeled using a Gaussian Process (GP) in BO.

The key advantage of EI is its preference for regions with high improvement potential. This makes it particularly useful for hyperparameter optimization where computational resources are limited, as it directs the search towards areas most likely to yield beneficial results. In the context of model fusion for language models, applying EI to the outer optimization loop of the bilevel optimization framework encourages exploration of hyperparameter settings.

\subsection{Upper Confidence Bound (UCB)}

\noindent Upper Confidence Bound (UCB) is another widely adopted acquisition function in Bayesian Optimization, with a focus on exploration. UCB emphasizes regions of high uncertainty.

The UCB acquisition function is given by: \cite{pesmo}
\[
UCB(\theta) = \mu(\theta) + \kappa \sigma(\theta),
\]
where \( \mu(\theta) \) represents the predicted mean of the objective at point \( \theta \), \( \sigma(\theta) \) is the predicted standard deviation (uncertainty), and \( \kappa \) is a parameter controlling the exploration-exploitation trade-off. A larger value of \( \kappa \) encourages more exploration by giving greater weight to the uncertainty term, while smaller values of \( \kappa \) favor exploitation, where the search focuses on regions with high predicted mean values.
UCB is suitable for the inner loop of bilevel BO to explore the parameter space for training configurations that may be uncertain but could offer significant improvements. By targeting uncertain regions, it prevents premature convergence and ensures a thorough search for optimal parameters.

\subsection{Bilevel Optimization Framework} 
\noindent Bilevel optimization provides a principled approach to nested optimization problems commonly encountered in machine learning. The general formulation is:

\begin{equation} \min_{\theta} F\bigl(\theta, \phi^{}(\theta)\bigr) \quad \text{subject to} \quad \phi^{}(\theta) = \arg \min_{\phi} G(\theta, \phi), \label{eq:bilevel} \end{equation}

\noindent where $\theta$ represents the \emph{outer} parameters (typically hyperparameters) and $\phi$ the \emph{inner} parameters (typically model parameters). The outer objective $F$ evaluates validation performance or other high-level criteria, while the inner objective $G$ corresponds to training loss minimization.

\noindent In our implementation, the bilevel framework operates as follows: \begin{enumerate} \item The outer loop maintains a probabilistic model (Gaussian Process) of the validation performance landscape \item An acquisition function selects promising hyperparameter configurations $\theta$ \item For each $\theta$, the inner loop optimizes model parameters $\phi$ to minimize $G(\theta,\phi)$ \item The resulting $\phi^*(\theta)$ is evaluated on validation data to update $F$ \end{enumerate}

This decoupling allows specialized strategies for each level: the outer loop focuses on generalization performance while the inner loop concentrates on training optimization. The Bayesian perspective comes from modeling $F$ as a Gaussian Process, enabling principled uncertainty estimation to guide the search.

\section{Optimizing Model Fusion Through Acquisition Function Design}
\begin{figure}[htbp]
    \centering
    % 第一行并排显示三图
    \begin{minipage}[t]{0.32\textwidth}  % 调整宽度比例为1/3左右
        \centering
        \includegraphics[width=\linewidth]{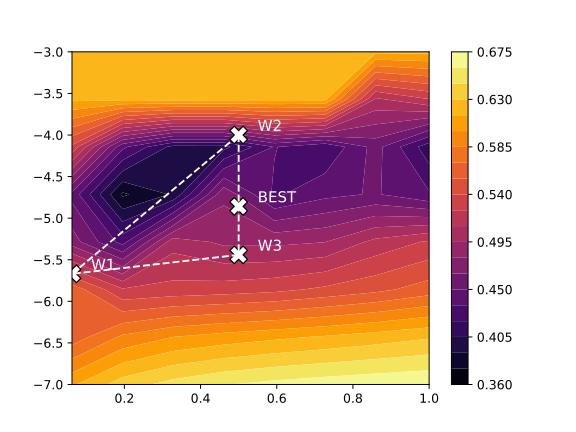} % 使用minipage的宽度
        \subcaption{}
    \end{minipage}
    \hfill % 填充水平间距
    \begin{minipage}[t]{0.32\textwidth}
        \centering
        \includegraphics[width=\linewidth]{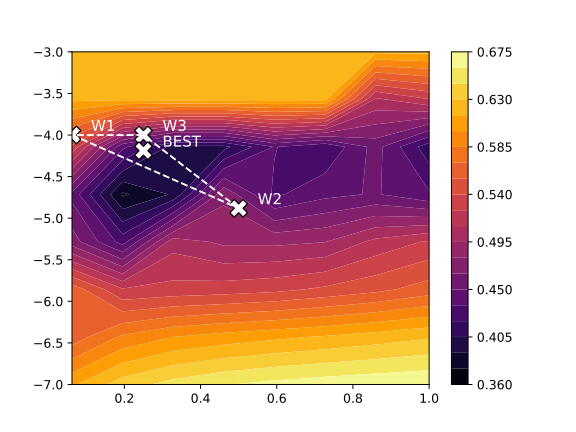}
        \subcaption{}
    \end{minipage}
    \hfill
    \begin{minipage}[t]{0.32\textwidth}
        \centering
        \includegraphics[width=\linewidth]{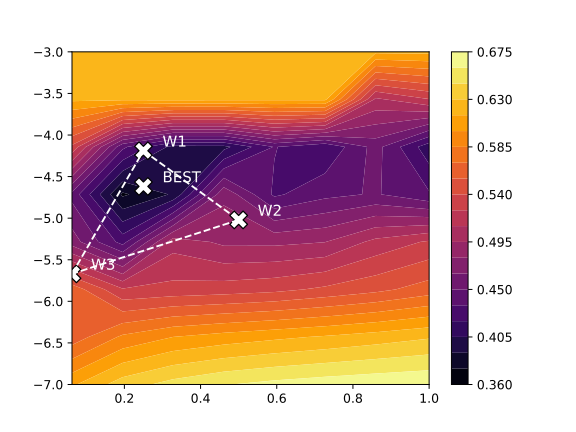}
        \subcaption{}
    \end{minipage}
    \caption{The figure illustrates the search spaces of different acquisition functions in our method.}
\end{figure}

 \noindent In our work, The bilevel Bayesian Optimization (BO) framework is extended by focusing on the optimization of acquisition functions, which play a pivotal role in guiding the search process. To this end, we systematically explored combinations of Expected Improvement (EI) and Upper Confidence Bound (UCB) both separately and together. Specifically, we evaluated scenarios where EI was employed in the outer optimization layer and UCB in the inner layer, as well as the inverse configuration. These experiments were designed to assess how different pairings of acquisition functions influence the identification of optimal model fusion parameters. Our findings demonstrate that the choice and arrangement of acquisition functions significantly impact the generalization performance, highlighting the necessity of tailored acquisition function design in bilevel optimization for large language models.

 The figure 1 shows the detailed search space in different acquisition functions. From left to right are: the inner layer using UCB with the outer layer using EI; both layers using EI; and the inner layer using EI with the outer layer using UCB. The last approach demonstrates the best performance, and its search space is notably larger than those of the other two methods.
\section{Experiment setup}
\subsection{Dataset and BenchMark}
\noindent In our experiments, we evaluate Bayesian‐optimized RoBERTa‑base on the General Language Understanding Evaluation (GLUE) benchmark, a widely adopted suite of nine natural language understanding tasks designed to probe diverse linguistic phenomena. GLUE comprises both single‐sentence (e.g., CoLA for acceptability judgment, SST‑2 for sentiment classification) and sentence‐pair (e.g., MNLI for multi‑genre NLI, QQP for paraphrase identification) tasks drawn from varied corpora. Each task supplies its own training, development, and test splits, with evaluation carried out using task‑specific metrics such as Matthews correlation (CoLA), accuracy (SST‑2, MNLI), or F1 score (QQP). By aggregating performance into a unified score, GLUE provides a rigorous means to compare different model configurations under a common evaluation protocol, making it an ideal testbed to assess the impact of Bayesian hyperparameter search on RoBERTa-base's generalization capabilities. \cite{wang2018glue}

\subsection{Model Architecture}
\noindent We adopt the RoBERTa‑base encoder \cite{liu2019roberta}, comprising 12 Transformer layers, each with hidden size 768 and 12 self‑attention heads. The model is initialized from HuggingFace’s pre‑trained weights, and all layers remain trainable during fine‑tuning. We apply layer‑norm after each sub‑layer and use GELU activations throughout.  

\subsection{Hyperparameter Settings}
\noindent Bayesian optimization searches over four hyperparameters:  
\begin{itemize}
  \item \textbf{Learning rate} (log–uniform in $[1\times10^{-6},\,1\times10^{-5}]$)  
  \item \textbf{Batch size} (\{8, 32\})  
  \item \textbf{Weight decay} (uniform in $[0.0,\,0.1]$)  
\end{itemize}
We use a Gaussian Process surrogate with Expected Improvement as the acquisition function, running 50 trials. Each trial fine‑tunes for up to 10 epochs with early stopping (patience=3) on the development set. All runs use a fixed random seed; when no official dev split exists, we hold out 10\% of training data.  

\section{Experiment}
\begin{figure}[htbp]
\centering
\includegraphics[width=120mm]{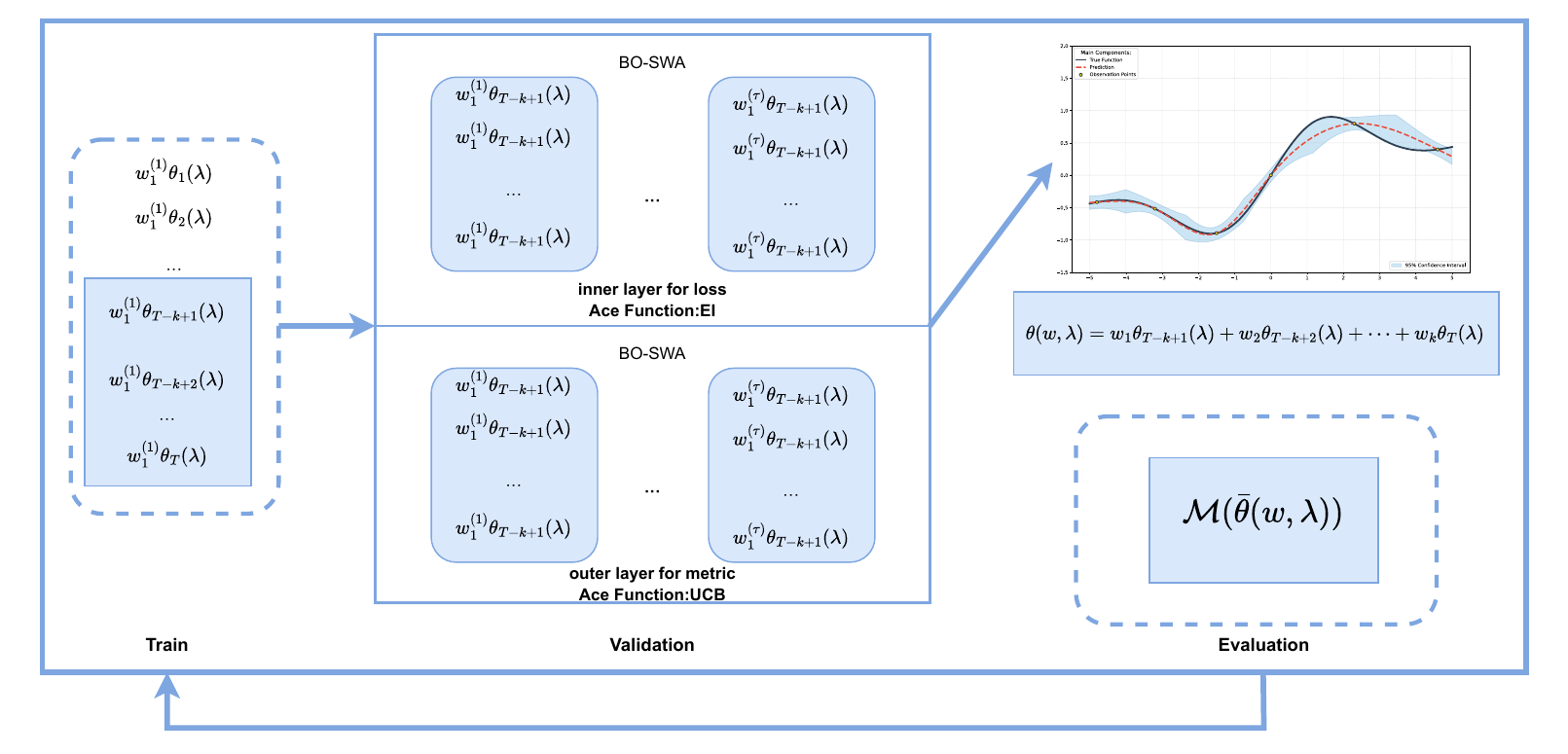}
\caption{Schematic illustration of the Bilevel Bayesian Optimization framework.}
\label{fig2:env}
\end{figure}
% Requires: \usepackage{graphicx}

\noindent In this section, we detail the experimental procedure and present the corresponding results. Fine-tuning is employed as a baseline to benchmark the performance of our proposed bilevel Bayesian optimization (BO) framework. Fine-tuning directly optimizes all model parameters on the target dataset, without utilizing a hierarchical optimization strategy. For our bilevel BO framework, we evaluate multiple acquisition function configurations \cite{bo_llm}, including: (1) Expected Improvement (EI) for both the inner and outer optimization loops, (2) Upper Confidence Bound (UCB) for both loops, (3) EI in the inner loop and UCB in the outer loop (referred to as "EI-UCB"), and (4) UCB in the inner loop and EI in the outer loop ("UCB-EI").

The bilevel coordination enables joint optimization of parameter configuration and ensemble composition, where the inner layer refines model capacity while the outer layer strategically combines model variants to enhance generalization performance. Distinct acquisition functions are deliberately designed for their respective optimization objectives, establishing synergistic interaction between the two levels.
\begin{table}[h!]
    \caption{GLUE results using RoBERTa-base. Bold indicates best performance. Improving Rate is relative to Fine-tune's AVG.}
    \centering
    \small  % 缩小字体
    \begin{tabular}{@{}lcccccc@{}}
        \toprule
        Method (In-Out) & RTE & MRPC & CoLA & STS-B & AVG & Imp. Rate (\%) \\ 
        \hline
        \midrule
        Fine-tune       & \textbf{71.2} & 88.4 & 55.6 & 84.0 & 74.80 & -- \\
        \hline
        single-level	& 71.2 & 88.4 & 60.3 & 84.0 & 75.98 & 1.18
        \\
        \hline
        UCB{(Ours)}       & 70.9 & 89.0 & 54.3 & 87.0 & 75.30 & 0.67 \\
        \hline
        EI(Ours)        & 70.8 & 88.5 & 58.8 & 84.0 & 75.52 & 0.96 \\
        \hline
        UCB-EI(Ours)     & 70.8 & 89.7 & 54.3 & 87.0 & 75.45 & 0.87 \\
        \hline
        EI-UCB{(Ours)}     & 70.8 & \textbf{90.6} & \textbf{58.8} & \textbf{87.1} & \textbf{76.82} & \textbf{2.70} \\
        \bottomrule
    \end{tabular}
    \label{tab:performance_comparison}
\end{table}
\FloatBarrier  % 强制后续内容出现在表格之后
Table 1 presents the GLUE benchmark results, highlighting the performance of different acquisition function combinations within our bilevel BO framework. Among the configurations, EI-UCB achieves the highest average score (76.82), outperforming standard fine-tuning by 2.7\%. These results strongly validate the effectiveness of our proposed optimization strategy and underscore its contribution to advancing model fusion techniques.
% Requires: \usepackage{graphicx}

\begin{table}[h!]
    \centering
    \caption{Loss performance comparison across different tasks. Bold indicates best performance.}
    \begin{tabular}{@{}lcccccc@{}}
        \toprule
        {Method(In-Out)} & {RTE} & {MRPC} & {CoLA} & {STS-B} & {AVG} \\
         \hline
        \midrule
        UCB(Ours) & 1.42 & 0.69 & 0.74 & \textbf{0.42} & {0.81} \\
        \hline
        EI(Ours) & 1.41 & 0.66 & 0.68 & 0.44 & 0.79 \\
        \hline
        UCB-EI(Ours) & 1.41 & 0.56 & 0.74 & 0.45 & 0.79 \\
        \hline
        EI-UCB(Ours) & \textbf{1.41} & \textbf{0.55} & \textbf{0.68} & 0.43 & \textbf{0.76} \\
        \bottomrule
    \end{tabular}
    \label{tab:performance_comparison}
\end{table}
Table 2 presents the loss performance for four tasks under different configurations of our approach. Ours(EI-UCB) achieves the lowest average loss (0.76) and also yields the minimal MRPC loss (0.55), demonstrating its overall superiority in loss minimization. Meanwhile, Ours(UCB) attains the smallest STS-B loss (0.42), and Ours(EI) ties with Ours(EI-UCB) for the lowest CoLA loss (0.68). These findings highlight the effectiveness of the EI-UCB strategy in consistently achieving robust performance across multiple tasks.

\begin{figure}[htbp]
    \centering
    % 并排显示两图
    \begin{minipage}[t]{0.48\textwidth}  % 调整宽度比例为接近1/2
        \centering
        \includegraphics[width=\linewidth]{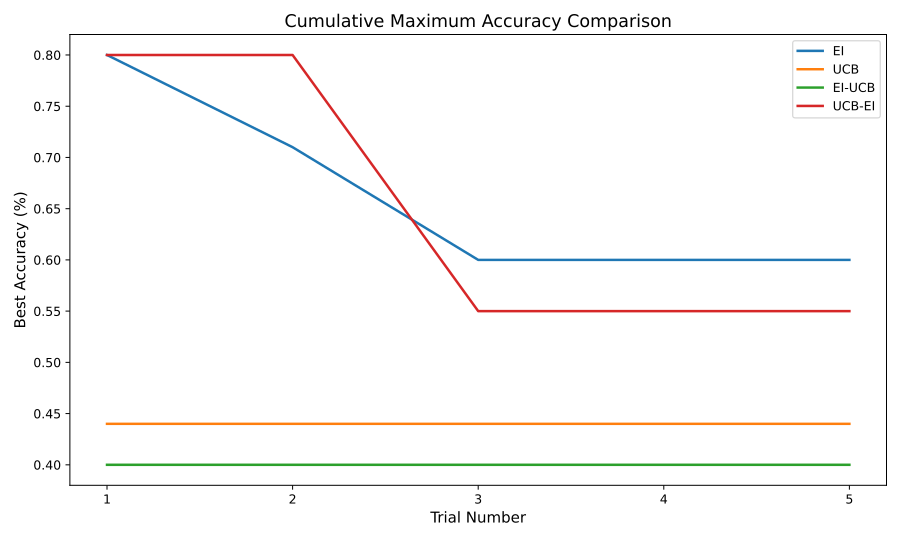} % 替换为你的图片路径
        \subcaption{}
    \end{minipage}
    \hfill % 填充水平间距
    \begin{minipage}[t]{0.48\textwidth}
        \centering
        \includegraphics[width=\linewidth]{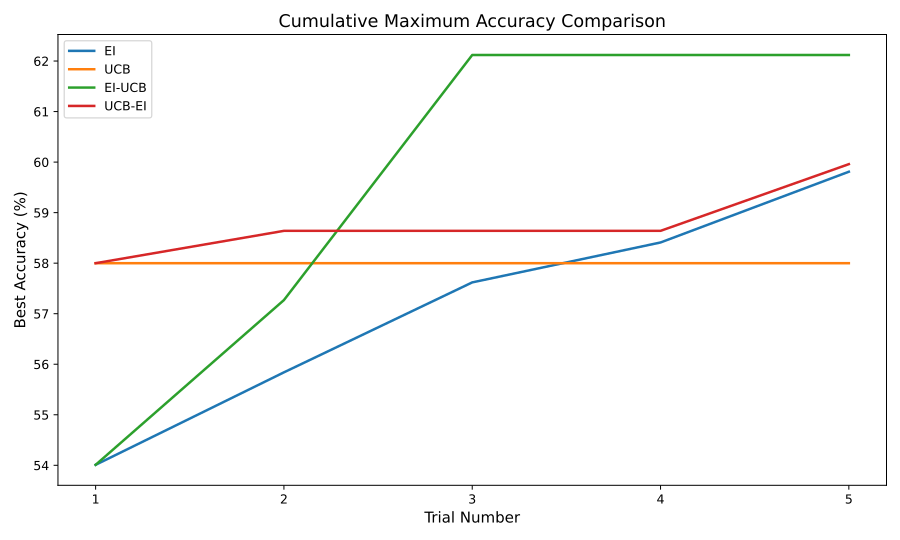}
        \subcaption{}
    \end{minipage}
    \caption{Cumulative Maximum Accuracy Comparison.}
\end{figure}

\noindent Figure 3 illustrates the performance of different acquisition function combinations by tracking the loss and accuracy (acc) metrics across multiple trials. Our optimal combination demonstrates efficient convergence, achieving the best results (minimized loss and maximized accuracy) within a limited number of trials.

\section{Limitations and future work}

\noindent Although the proposed bilevel optimization framework shows better performance for hyperparameter tuning on language model fine tuning, some limitations need to be discussed.

\subsection{Model Generalizability}

\noindent We only test our experiments on RoBERTa, which raises the question as to whether or not this works for other transformers, such as big ones with different internal logic. This is short of doing a big study where we check if it works well on different types of computers too, which is something we should do later on.

\subsection{Computational Efficiency}

\noindent The second point is computational efficiency. Though our two-layer optimization framework provides significant speedup compared to grid search on the optimization of a large number of hyperparameters, it still suffers from high computational cost. And this method really works well if a lot of those things are adjusted at once. However, if only a few core hyper-parameters are needed (e.g. only learning rate or weight decay), the overhead of keeping a GP model up to date will be too costly compared to the benefit \cite{shahriari2016taking}. In this setting traditional mesh search is more accurate and efficient. Adaptive strategies to switch between the two methods according to the size and complexity of the hyperparameter space based on the combination of the two methods.

\subsection{Acquisition Function Selection}

\noindent The current implementation's use of the EI-UCB combined acquisition function is effective based on our experiments but is still a limited exploration of the space of possible designs. Alternative methods worth investigating include probabilistic combinations of acquisition functions \cite{gonzalez2016predictive} (such as Thompson Sampling with Probability of Improvement), dynamically switching based on optimization progress, and multifidelity evaluation strategies. Theoretically, it seems that different acquisition functions may be better depending on the problem, so an exploration focused method such as UCB might work when improvement focused methods fail under constrained resources. Future work can study these relationships more systematically and come up with principled function-selection criteria.

\section{Conclusion}

\noindent In this work, {Bilevel-BO-SWA}, a novel approach that unifies model fusion with a bilevel optimization strategy to enhance large language model fine-tuning. By systematically exploring and pairing acquisition functions (EI and UCB) within the inner and outer loops, our method achieves better generalization over traditional fine-tuning, and prior fusion strategies on GLUE tasks.

\section*{Acknowledgement}
\noindent This preprint has no post-submission improvements or corrections. The Version of Record of this contribution is published in the Neural Information Processing, ICONIP 2025 Proceedingsand is available online at https://doi.org

\end{document}